\documentclass[letterpaper, 10 pt, conference]{ieeeconf}  

\IEEEoverridecommandlockouts                              
\usepackage{cite}
\usepackage{amsmath,amssymb,amsfonts}
\usepackage{algorithmic}
\usepackage{graphicx}
\usepackage{textcomp}
\usepackage{math}
\usepackage{multirow}
\usepackage{color}
\newcommand{\comment}[1]{}
\newcommand{\rev}[1]{{#1}}

\begin{document}
\title{\LARGE \bf
SUPR-GAN: SUrgical PRediction GAN for Event Anticipation\\
in Laparoscopic and Robotic Surgery
}

\author{Yutong Ban$^{1,2}$, 
Guy Rosman$^{1,2}$,
Jennifer A. Eckhoff$^{2}$, Thomas M. Ward$^{2}$, Daniel A. Hashimoto$^{2}$\\
Taisei Kondo$^{3}$, 
Hidekazu Iwaki$^{3}$, 
Ozanan R. Meireles$^{2}$ and 
Daniela Rus$^{1}$
\thanks{$^{1}$Distributed Robotics Laboratory, CSAIL, MIT. 32 Vassar St, Cambridge, MA, US
        {\tt\small \{yban, rosman, rus@csail.mit.edu\} }}
\thanks{$^{2}$SAIIL, Massachusetts General Hospital. 55 Fruit Street, Boston, MA, US 
        {\tt\small \{jeckhoff, tmward, dahashimoto, ozmeireles@mgh.harvard.edu\}}}%
\thanks{$^{3}$Olympus Corporation, Tokyo, Japan{\tt\small \{taisei.kondo2, hidekazu.iwaki@olympus.com\}}}
\thanks{This work is supported by the research fund from Olympus Corporation.}
}

\markboth{IEEE Robotics and Automation Letters. Preprint Version. Accepted February, 2022.}
{Ban \MakeLowercase{\textit{et al.}}: SUPR-GAN: SUrgical PRediction GAN for Event Anticipation in Laparoscopic and Robotic Surgery} 

\maketitle

\begin{abstract}
    Comprehension of surgical workflow is the foundation upon which artificial intelligence (AI) and machine learning (ML) holds the potential to assist intraoperative decision making and risk mitigation. In this work, we move beyond mere identification of past surgical phases, into prediction of future surgical steps and specification of the transitions between them. We use a novel Generative Adversarial Network (GAN) formulation to sample future surgical phases trajectories conditioned on past video frames from laparoscopic cholecystectomy (LC) videos and compare it to state-of-the-art approaches for surgical video analysis and alternative prediction methods. We demonstrate the GAN formulation's effectiveness through inferring and predicting the progress of LC videos. We quantify the horizon-accuracy trade-off and explored average performance, as well as the performance on the more challenging, and clinically relevant transitions between phases. Furthermore, we conduct a survey, asking 16 surgeons of different specialties and educational levels to qualitatively evaluate predicted surgery phases. 
\end{abstract}

\section{Introduction}
Surgical artificial intelligence (AI) ultimately aims to create machines capable of improving patient care.
Whilst surgical AI and machine learning (ML) research advances, major focus has been placed on building the foundation for machines to comprehend surgical workflow\cite{lalys2014surgical,volkov2017machine,twinanda2016endonet,jin2017sv, ward2021computer, ward2021challenges, ward2022artificial, meireles2021sages}. Investigations mostly concentrate on post-hoc analysis, trying to identify the current surgical phase from past video events alone. 
However, in order to foresee and prevent surgical complications, machines must be able to predict future events. Previous research has neglected prediction of future events or operative phases, only briefly reporting on specific predictive tasks, such as remaining surgery time \cite{twinanda2018rsdnet} and tool anticipation \cite{rivoir2020rethinking}.
The ability to predict multiple, non-sequential, future possible phases is especially prominent in more complex surgical cases where differences in subsequent phases could lead to drastically different outcomes. Data-driven modeling of the surgical workflow and automatic methods of risk prediction during such procedures could yield a tremendous benefit for both the patient and surgeon. With the advances in robotic surgery and synergistic integration of computer vision, AI augmented surgery may in the far future translate the potential of advanced driving assistance systems to the operating room \cite{hashimoto2018artificial}. Phase prediction and early hazard detection promises great potential for the improvement of patient outcomes, especially when paired with  surgical robotic innovation (\cite{Hashimoto2021-yb}, ch. 12.)

In order to provide more general ability to predict future events during ongoing surgery, we need more capable models that truly understand surgical workflow. 
Such models should be able to handle multiple inference tasks and account for the complexity and actions throughout a surgical procedure. They should comprehend discrete characteristics of surgical procedures, such as abrupt transitions between phases, surgeons' decisions, and transitions in clinically-meaningful variables.
Their predictive capability should extend beyond the mere ability to predict specific narrow tasks such as remaining surgical time, accounting for the diversity of phenomena that occurs during surgery. 
Finally, they should handle  the uncertain nature of the sequences involved,  both due to the limited field-of-view and the third party observer's lack of theory-of-mind of surgeon's intent and actions context,  as well as the inherent uncertainty of human decision making -- we expect predictive models to afford probabilistic reasoning that robustly handles such phenomena. 

\begin{figure}[t!]
    \centering
    \includegraphics[width=.48\textwidth]{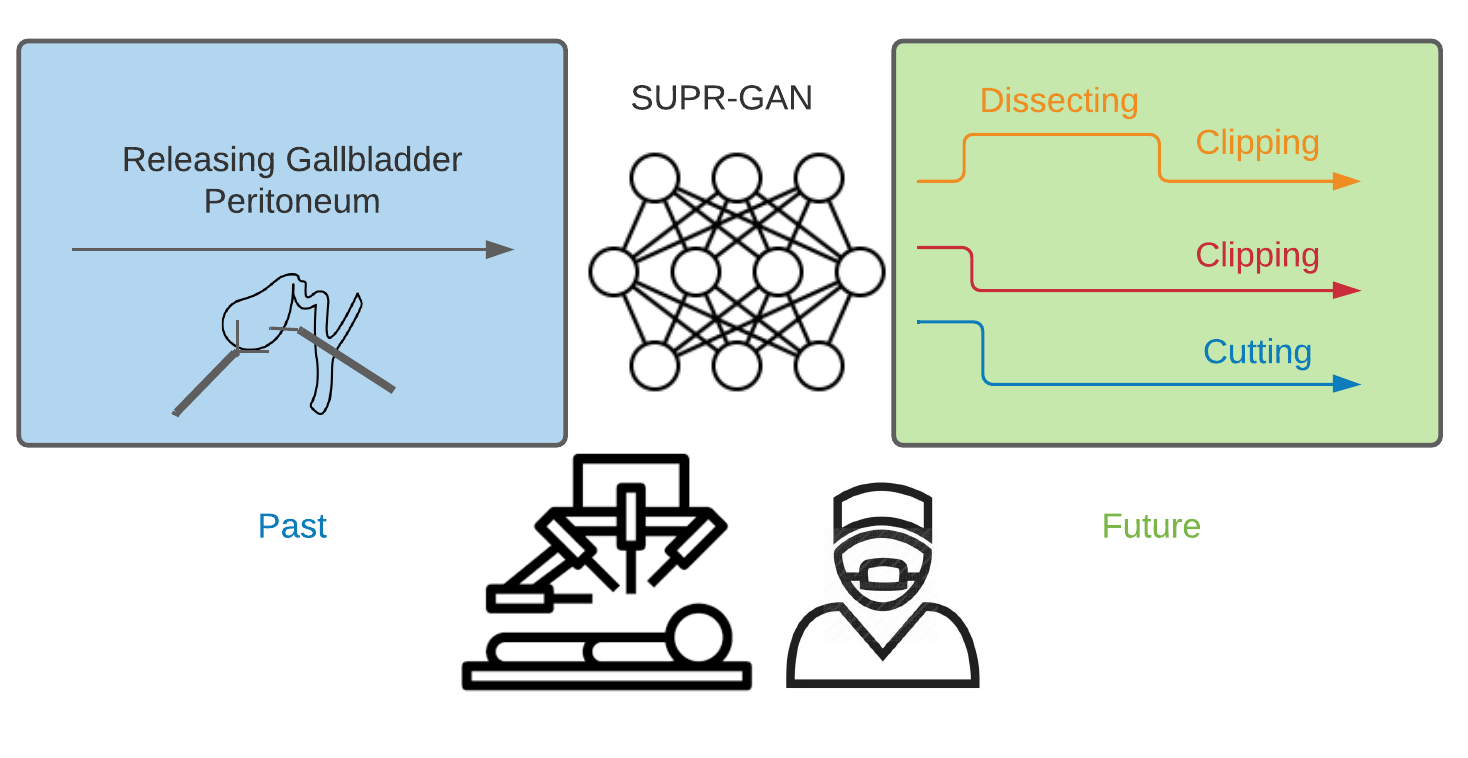}
    \caption{Model overview: we predict a distribution of alternative future surgical phase sequences \rev{based on past surgical video frames}.}
    \label{fig:teaser}
    \vspace{-3ex}
\end{figure}

In this work, we focus on surgical phase prediction in laparoscopic cholecystectomy (LC) - removal of the gallbladder. With over 500,000 laparoscopic and robotic cases performed in the US annually, the cholecystectomy is among the most frequent surgical procedures. Although routinely performed, common bile duct injuries (with an incidence of up to $ 3\%$ in LC) present a serious and frequent complication with major, potentially life threatening consequences for the patient \cite{renz2017bile}. Therefore, early and advanced recognition of undesirable deviations from the routine surgical course with prediction and avoidance of adverse events like bile duct injuries holds great risk mitigation potential and can improve patient outcome significantly. 
\rev{The prediction horizon should be sufficiently long to provide time to take action in order to prevent complications resulting from the immediate actions of the surgeon, such as accidental injury of the common bile duct or right hepatic artery. In response to consultation of a panel of board certified surgeons, a prediction horizon of 15 seconds was chosen for this work. This  provides sufficient time to foresee future steps and potentially recruit further expertise and assistance in complicated cases. A longer prediction horizon may distract the surgeon from the standardized surgical workflow.}

We address this prediction problem via a novel model that uses an encoder-decoder predictor based on a discrete generative adversarial network (GAN \cite{goodfellow2014generative}). This model\rev{, illustrated in Figure~\ref{fig:teaser},} can predict roll-outs of the surgical process in the form of discrete label sequences for operative phases over time\rev{, by observing the recent laparoscopic video frames}. Also, the model emits the phase estimates over the past and current time phase, allowing analysis and prediction of the surgery within a unified, multitask framework.

\noindent\textbf{Contributions:} Our contributions are:
\begin{enumerate}
    \item We define and explore a novel, yet important problem of surgical prediction and jointly estimate the phase analysis.
    \item We examine models for surgical process prediction, including a new discrete GAN predictor - SUrgical  PRediction GAN (SUPR-GAN). 
    \item We demonstrate results of the proposed model for analysis and prediction of LC. We show how the model surpasses existing approaches in both sequence prediction and transition detection tasks. We further quantify the subjective plausibility of the predictions based on a survey of surgeons.
\end{enumerate}

\section{Related works} 
Our work relates to several major topics in current AI/ML and surgical prediction research. Significant effort has been made in surgical workflow analysis. \cite{twinanda2016endonet} was one of the first to propose deep-learning architecture to address the problem by applying a convolutional neural network (CNN). \cite{jin2017sv} first introduced CNN-LSTMs for this task, nowadays considered standard. Additional research explored the applications to more complicated procedures, such as sleeve gastrectomy in \cite{volkov2017machine} and to the peroral endoscopic myotomy (POEM)  in \cite{ward2020automated}. Moreover, due to the potentially lengthy duration of the procedure, \cite{ban2021aggregating} proposed to aggregate the long-term temporal dependencies with statistic features so that the model can capture global surgical information. Similarly, \cite{jin2021temporal} proposed a temporal memory relation network to capture multi-scale temporal patterns.

Aside from phase recognition, tool and task identification \cite{twinanda2016endonet,nwoye2020recognition} is another active field in the surgical AI domain. A CNN was successfully applied to capsule endoscopy to detect endoluminal lesions in the bowel \cite{szegedy2016rethinking} and colour descriptors have been used to assess intraoperative bleeding \cite{garcia2017automatic}. \cite{uemura2018feasibility} applied a neural network classifier to evaluate kinematic data from robotic surgery and distinguish between more and less experienced surgeons. However, little effort has been devoted to prediction of future workflow with a few notable exceptions, such as prediction of remaining surgical time as a regression task \cite{twinanda2018rsdnet}, or the relatively unimodal next action \cite{Park2021-pr}. Retrospective automated analysis of intraoperative adverse events (bleeding, cautery injury) has been performed, emphasising the significant impact of such events in terms of increased morbidity, mortality, and hospital stay \cite{wei2021intraoperative}. Whilst these adverse events most commonly arise from preventable errors, little research has been conducted on the prospective prediction of subsequent intraoperative phases and events/errors to enhance surgical outcome. 

In context-aware systems in computer assisted interventions (CA-CAI), using both real-time visual and instrumental information has been shown to increase accuracy in surgical workflow identification \cite{dergachyova2016automatic}, aiding OR situational awareness. Teaching surgical trainees technical skills, comprehension of surgical workflow, and decision making in a protected setting (without potential complications to the patient) holds great value to improve clinical education for surgeons. As shown by \cite{higgins2021development} simulation-based, video based learning and retrospective action analyses significantly improve technical as well as procedural surgical skills.

\rev{Sequence and trajectory prediction has been commonly applied to autonomous driving, where forecasting of road users' future actions allows vehicles to interact, plan and warn the driver of road risk \cite{huang2021carpal}. However, the trajectories predicted are in a continuous state space.  Other concerns, such as multi-agent interactions, agent goals, and environmental context, lead to different design structures \cite{lee2017desire,gupta2018social,rhinehart2019precog,ivanovic2020multimodal}. Moreover, GANs can be found in other prediction applications, such as motion prediction \cite{barsoum2018hp,kundu2019bihmp}, body-pose prediction\cite{zhao2020masked}, speech signal prediction \cite{juvela2019gelp}. However, they often merely predict continuous signals and hence limit the discrete aspect to either a discrete vector \cite{ivanovic2020multimodal} or a mixture model \cite{deo2018multi,huang2019uncertainty}, save for a few exceptions \cite{Huang2022icra,Li2022-wi}.   } 

Auto-encoding and processing of discrete sequences, with different types of connectivity structures and tasks, has been prevalent in the natural language processing (NLP) literature \cite{sutskever2014sequence,cho2014properties,dai2015semi,tai2015improved,bowman2016generating}. Outside of NLP, prediction and completion of discrete sequences has been more limited in its applications, with notable examples in several fields of natural sciences, such as chemistry and biology. \cite{kadurin2017cornucopia,yu2017seqgan,gomez2018automatic,costello2019hallucinate,yang2019machine}. However, in these domains, often the signal of interest is directly observed, unlike surgical phases that are not directly observed in the video.

\begin{figure*}[ht!]
    \centering
    \includegraphics[width=.85\textwidth]{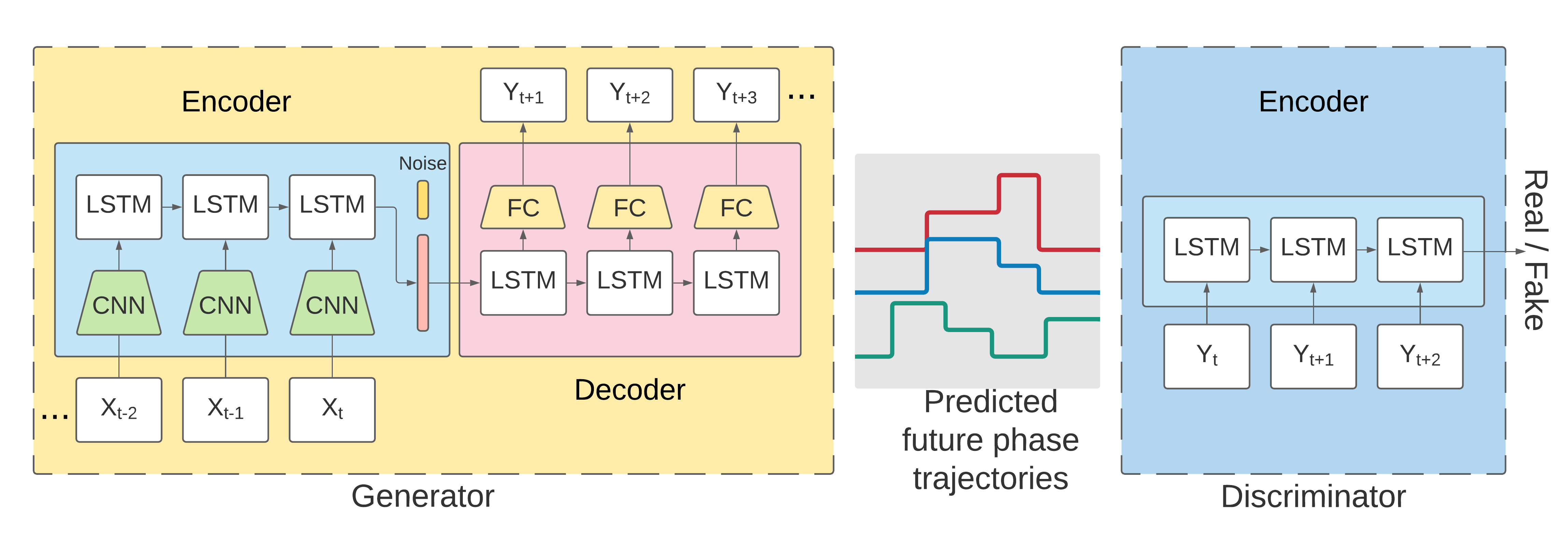}
    \caption{Overview of the proposed SUPR-GAN model. Our model includes a CNN-LSTM-based encoder and a discrete decoder.}
    \label{fig:my_label}
    \vspace{-3ex}
\end{figure*}

\rev{Overall GAN approach has several advantages in a surgical context: (i) GAN training w/ variety loss encourages diverse predictions \cite{gupta2018social}, that are important for workflow prediction at decision-making junction during surgical procedures. (ii) The GAN model can be easily modified for discrete sequences \cite{kusner2016gans} as surgical phases and intra-operative events are often discrete labels. (iii) It extends gracefully to the prediction of multiple complex entities, as needed to explain surgical processes at adverse events. Therefore we consider GAN to be the most appropriate framework to address this surgical prediction question.}


\section{Methods}

In this section we elaborate the research question and the detailed design of the proposed prediction model. 
\subsection{Problem formulation}
\label{sec:problem formulation}
A surgeon can currently be performing one of $N_{P}$ possible phases during an operation. Our goal is to predict the surgeon's future actions. Instead of predicting the only possible phase trajectory, we try to predict a distribution of possible phase trajectories using GAN, limiting the prediction learning to a fixed past and future horizon, as is common in prediction literature. During surgery, a surgeon makes decisions based on the recently observed $T_{p}$  video frames $\Imat = \{\Imat_{t_0-T_{p}}, \dots, \Imat_{t_0-1}, \Imat_{t_0}$\}. The model prediction is represented by $\ymat = \{\ymat_{t_0+1},  \ymat_{t_0+2}, \dots \ymat_{t_0+T_{f}}$\}, where $T_{f}$ is the number of frames the model is predicting into the future. At each time step, the model predicts the surgical phases $\ymat_{t} \in \{0,1\}^{N_{P}}$, where phase labels are encoded as 1-hot vectors.

\subsection{Auto-encoding Sequence Models}
Several auto-encoding models have been used to predict sequential data, mostly based on GANs \cite{goodfellow2014generative,gupta2018social,huang2020diversitygan} and conditional variational auto-encoders \cite{sohn2015learning,ivanovic2020multimodal}.
Our model follows GANs and includes a generator and a discriminator model denoted by $\mathbf{G}$ and $\mathbf{D}$ respectively.  The generator includes an encoder and a decoder. The discriminator includes a past encoder and a future encoder. It is trained to distinguish whether a presented trajectory is a fake trajectory created by the generator, or whether it is a real trajectory from the data. Usually, an additional data term is added to ensure that the generator can regenerate real data trajectories, often with a variety loss \cite{gupta2018social,thiede2019analyzing}.
\subsection{Surgical Prediction GAN (SUPR-GAN)}
Unlike standard sequence prediction problems, we have additional supervisory cues in the form of annotation labels, $\hat{\Ymat}$.
The overview of the network is given in Fig. \ref{fig:my_label}.


\noindent \textbf{Generator Encoder} The generator's encoder utilizes the observations and encodes all the observed information into a single vector. Since the observations are sequential, long short-term memory (LSTM) is thus applied for the recurrence. The encoding process can be formally written as:
\begin{align}
\evect_{I} &= CNN(\Imat_{t}) \nonumber \\ \hvect_{t} &= LSTM_{G,\textbf{enc}}(\hvect_{t-1}, \evect_{I})\\  
\ymat_{t} &= PhaseHead(\hvect_{t}) \nonumber
\end{align}
Where CNN uses ResNet\cite{he2016deep} as the backbone, and $\hvect$ represents the hidden state of the LSTM. Both the encoder and the decoder hidden state share the same dimension $H = 32$. The $PhaseHead$ is a fully-connected layer which generates the phase probabilities.\\

\noindent \textbf{Generator Decoder} The decoder leverages the information from the encoder. While initializing the hidden state of the decoder, we simply take the last encoder hidden state then concatenate it with a random noise vector. A fully-connected layer is used to map the vector to size $H$. 
\begin{align}
\ymat_{t-1} = PhaseHead(\hvect_{t-1})  \\ \hvect_{t} = LSTM_{G,\textbf{dec}}(\hvect_{t-1}, \ymat_{t-1}) \nonumber
\end{align}
at each time step $t$, the estimated variables can be obtained by phase emission heads separately. 

\noindent \textbf{Discriminator}
The discriminator is composed of two encoders, one for each of the past and future sequences. Both encoders feed off phase vectors,
\begin{align}
    \hvect_{t} = LSTM_{D,\cdot}(\hvect_{t-1}, \ymat_{t-1}).
\end{align}
The last state of the future encoder is fed through a discriminator head that emits a binary label - real or fake - for the sequence.
The real represents the trajectory sampled from the real data, whereas fake represents the sample generated by the prediction model.   

\noindent\textbf{Discrete GAN} As the surgery workflow which we are predicting has discrete sequences, a function which can convert the generator decoder output probabilities to discrete sequences is required. Gumbel-Softmax \cite{jang2016categorical} layer is used after the generator decoder output. It is a differentiable layer so that it allows us to have discrete phase prediction samples input to the discriminator, without breaking the gradient. 

\noindent \textbf{Loss}
The loss function used for training is a combination of GAN loss and data loss. The GAN loss penalizes the whether the predicted trajectory is reasonable or not.
\begin{equation}
\mathcal{L}_{dis}(\ymat, \hat{\ymat}) = \min_{D} \max_{G} V(G,D),
\end{equation}
where $V(G,D)$ is the usual GAN loss, formally written as:
\begin{align}
    \min_{D} \max_{G} V(G,D) &=  \\
    \mathbf{E}_{x \sim p_{data}(x)} & \log(D(x)) + \mathbf{E}_{z \sim p_{z}(z)}( \log(1 -D(G(z))). \nonumber
\end{align}
$z$ is the generator noise sample and $x$ represent sample from the data's label sequence $\hat{\Ymat}$.
The data loss in GAN-based predictors is destined to ensure the prediction is not too far from the ground truth. We use a variety loss \cite{thiede2019analyzing}, which penalizes the distance between the ground truth labels and the most similar reconstructed sequence out of a set of $N_{s}=10$ samples.
\begin{equation}
\mathcal{L}_{\text{rec}}(\ymat, \hat{\ymat}) = \min_{j=1}^{N_s}\sum_{t=t_0+1}^{t_0+T_{f}} d_L(\ymat_{t}^{(j)}, \hat{\Ymat}_{t}).
\end{equation}
$d_L(\cdot,\cdot)$ is a distance between the labels and prediction -- cross-entropy as our sequences are discrete categories.



\noindent \textbf{Past Encoding Loss} Unlike domains where GAN-based predictors are decoding the raw signals (such as images and trajectory prediction), we are decoding annotated labels that are much more costly to obtain and are not the same as the encoded signal (images). This allows us to add an additional data term measuring how well the encoder recognizes the phase, even in past frames. This is similar to phase recognition costs and is expressed as:
\begin{equation}
\mathcal{L}_{\text{past}}(\ymat, \hat{\ymat}) = \sum_{t_0-T_p}^{t_0} d_L(\ymat_{t}, \hat{\ymat}_{t}),
\end{equation}
where we use cross-entropy loss as before.

During the experiment, the overall loss is
\begin{equation}
\mathcal{L} = \omega_1 \mathcal{L}_{dis} + \omega_2 \mathcal{L}_{rec} + \omega_3 \mathcal{L}_{past}
\end{equation}
where in practice $\omega_1 = 0.6$ and $\omega_2 = 0.2$ $\omega_3 = 0.2$.

\begin{figure*}[h!]
\centering
\includegraphics[width=.85\textwidth]{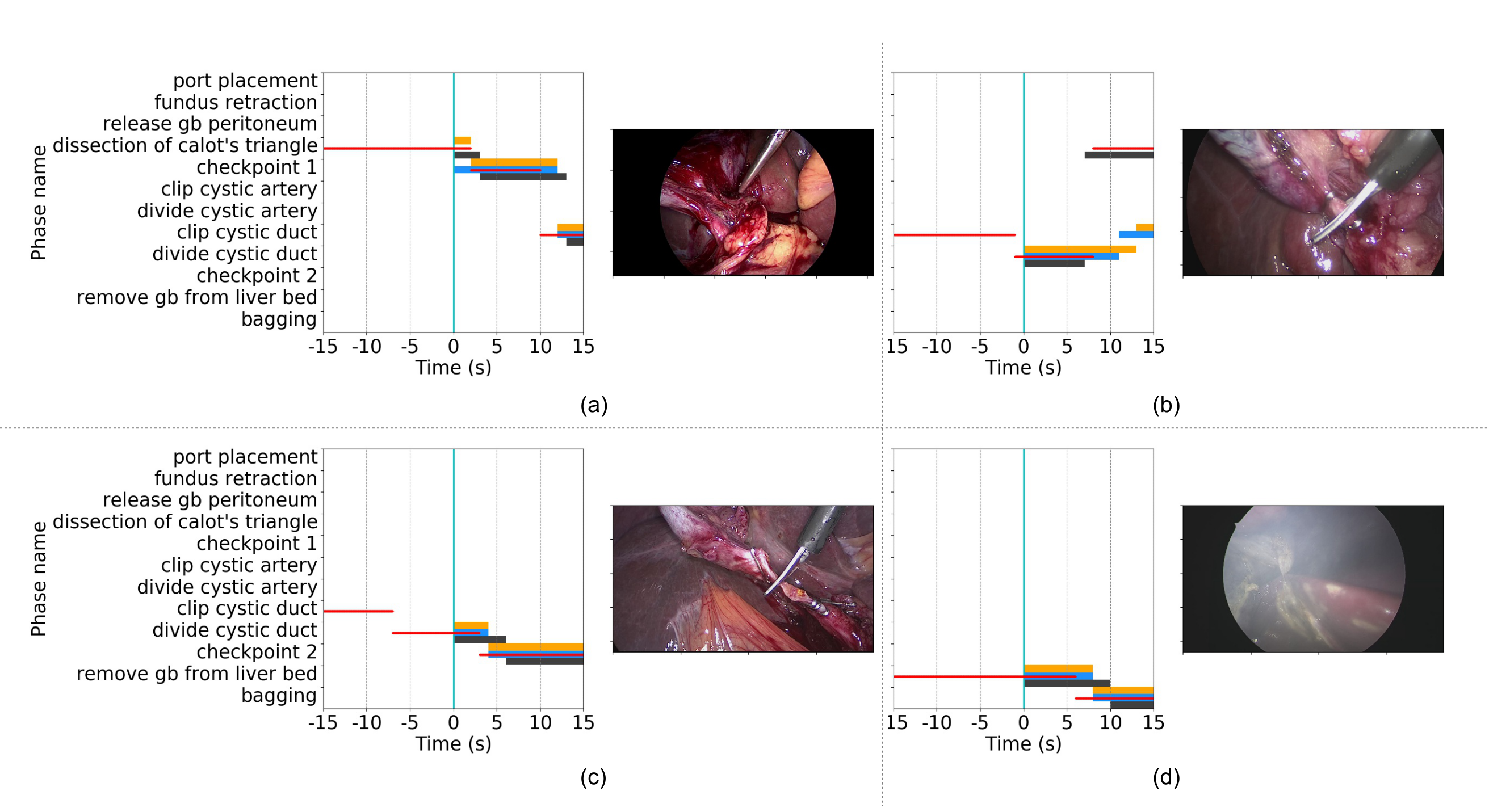}
\vspace{-1ex}
\caption{Examples of prediction results in MGH200 dataset. For each example, on the left, a diagram of the phases of the operation was shown. The horizontal axis indicates the time, ranging from the past 15s to future 15 seconds. The vertical cyan bar indicates the "current" time point associated with the video image on the right side. Red horizontal segments indicate the ground truth trajectories. And the horizontal color bars (orange, blue, black) indicate the different samples predicted by the SUPR-GAN, \rev{which are randomly chosen from the predicted samples.} (a) Video 20, frame 433 (b) video 41, frame 205 (c) video 20, frame 847, (d) video 47, frame 2535.}
\label{fig:qualitative results}
\end{figure*}

\begin{table*}
\centering
\begin{tabular}{c|c|c|c|c}
\hline
Transition to Phase & Constant Model & HMM  & Ours w/o Dis. & SUPR-GAN (Ours) \\
\hline
 Preparation &-&-& - &-\\
 Calot Triangle Dissection &7.5\%&5.3\% &72.5\% & 87.5\%\\
 Clipping and Cutting &2.5\%&42.5\%& 82.5\% &57.5\%\\
 Gallbladder Dissection &7.5\%&35\%& 70\%&65\%\\
 Gallbladder Packaging &12.5\%&10\%& 50\%&30\%\\
 Cleaning and Coagulation &42.8\%&31.4\%& 42.9\%& 54.3\%\\
 Gallbladder Retraction &85\%&17.5\%& 50\%& 30\%\\
\hline
Overall & 26.0\% & 31.5\%  & 59.6\% &\textbf{60.9\%}\\
\hline
\end{tabular}
\caption{The per-transition accuracy on Cholec80 Dataset. Our predictor significantly outperforms the baselines.}
\vspace{-7ex}
\label{tab:Cholec80}
\end{table*}

\section{Experiments}
\label{sec:experiments}
In the following we test both objective measures of accuracy relating to the model's predictive power, as well as the perceived plausibility of the predicted trajectories when gauged by clinical experts (surgeons).
\subsection{Datasets}
We experiment on two large-scale surgical video datasets.\\ \textit{Cholec80} \cite{twinanda2016endonet} is a publicly available dataset which contains 80 videos of LC. The dataset is divided into equal subsets for training and testing with 40 videos each. The dataset is annotated into 7 different phases. \\ \textit{MGH200} \cite{ban2021aggregating} consists of 200 LC videos, where 150 videos are used for training and 50 for testing. The dataset is annotated into 12 surgical phases, which is more granular than the Cholec80 datasets. MGH200 also contains more variability and clinically meaningful phase transitions.
\subsection{Model Parameters and Training Strategy}
In the experiment, the videos are re-sampled at 1 fps and fed to the model. During model training, the generator encoder is pre-trained in surgical phase recognition for 20 epochs. The pre-training is accomplished using the same dataset. Therefore, no additional data is used. During GAN training, we iteratively train the generator and discriminator, using small epochs, where the epoch size is 64 and the number of epochs is 2000. We used an Adam optimizer with a  learning rate of $10^{-4}$.

\subsection{Prediction settings}
Throughout prediction, we use the past 15 seconds to predict the upcoming future 15 seconds. This was determined for several reasons: (i) the past 15 seconds of the video segment should contain sufficient information about the operative phase to predict the future phase; (ii) a prediction horizon of 15 seconds into the future is an adequate time period for the surgeon to assess, intervene, and prevent  potential adverse events; (iii) LSTMs are limited in their ability to numerically propagate information over large timescales \cite{sutskever2013training} and insufficient in covering a complex set of predictions as the prediction horizon increases \cite{huang2020diversitygan,phan2020covernet}. In \ref{sec:ablation}, we also compared the effect of using different prediction horizons.

\subsection{Evaluation Metrics}
To evaluate the prediction models, we employ two metrics:\\
\vspace{-3ex}
\begin{itemize}{}
    \item \textit{Per-transition accuracy:} Every time the ground truth transits to a new phase, if said new phase is predicted correctly within $\delta$ seconds, we consider the transition to be well predicted. We set $\delta$ to 15 seconds. 
\item \textit{Levenshtein distance:} Levenshtein distance (LD) \cite{levenshtein1966binary} measures the minimum number of operations required to transform one sequence into another. It is widely applied to NLP for comparing strings and used to compare DNA sequences in biology \cite{al2017secure}. In our evaluation, we calculate the average Levenshtein distance between the prediction and the ground truth. 
\end{itemize}
\subsection{Qualitative Results}

The exemplary results of the proposed model are shown in Fig \ref{fig:qualitative results}. In example (a), the model predictions align with the ground truth during the transitions between 'Checkpoint 1' and 'Clip Cystic Duct'. Similarly, in the example (b), the transition from 'Clip Cystic Duct' to 'Dissection of Calot's Triangle' is well captured by the black prediction trajectory. Moreover, the model is able to detect the different possible future phases in more challenging and questionable phase transitions (e.g.in Figure \ref{fig:qualitative results} (b): the black trajectory shows the model is able to transit back from 'Clip the Cystic Duct' to further 'Dissection of the Calot's Triangle' as a reasonable surgical action). Other examples, confirming model and ground truth accordance, are shown in (c) and (d). The proposed model can not only give accurate prediction about the phase transitions, it can also predict alternative trajectories and therefore cover various possible transitions. 
\begin{table*}[t!]
\vspace{2ex}
\centering
\resizebox{.7\textwidth}{!}{
\begin{tabular}{c|c|c|c|c|c}
\hline
&Transition to Phase & Constant Model & HMM & Ours w/o Dis.  & SUPR-GAN(Ours) \\
\hline
 \multirow{5}{4em}{Block 1}& Port placement &0\%&0\%&0\%& 100\%\\
 &Fundus retraction &0\%&0\%&83.3\%& 83.3\%\\
 &Release GB peritoneum &91.8\%&32.1\%&67.9\%& 62.3\%\\
 &Dissection of Calot's triangle &12.3\%&69.5\%&61.9\%& 57.1\%\\
 &Checkpoint 1 & 0.0\%&100\%&33.3\%& 44.4\%\\
 \hline
 \multirow{5}{4em}{Block 2}&Clip Cystic Artery &1.8\%&12.7\%&7.3\%& 25.5\%\\
 &Clip Cystic Duct &2.1\%&8.5\%&29.8\%& 53.2\%\\
&Divide Cystic Artery &0\%&39.6\%&11.3\%&47.2\% \\
 &Divide Cystic Duct &0\%&0\%&60\%& 48.9\%\\
 &Checkpoint 2 &4.4\%&13.3\%&66.7\%& 57.8\%\\
 \hline
 \multirow{2}{4em}{Block 3} & Remove GB from liver bed &64.6\%&35.4\%& 81.9\%& 54.2\%\\
 &Bagging &91.8\% &100\%&83.8\%&86.8\%\\
 \hline
 & \textbf{Overall} &15\%&42\%&50.4\%& \textbf{53.5\%}\\
\hline
\end{tabular}
}
\caption{The per-transition accuracy on MGH200 Dataset. The SUPR-GAN out-performs the other methods in overall accuracy.}
\label{tab:mgh200}
\vspace{-5ex}
\end{table*}

\begin{table}[h!]
\centering
\resizebox{.45\textwidth}{!}{
\begin{tabular}{cc|c|c|c|c}
\hline
&Metrics & Constant  & HMM & Ours  &  SUPR-GAN   \\
& & Model &  &  w/o Dis.  &  (Ours) \\
\hline
  & LD (transitions) &9.53& 11.67& 9.26 &\textbf{9.15}\\
  & LD (Overall) & 3.47 & 13.15  &  4.27  &\textbf{3.36}\\
\hline
\end{tabular}
}
\caption{The Levenshtein distance (LD) on MGH200 Dataset (the lower the better). SUPR-GAN achieved the lowest LD in both overall and transition settings.} 
\label{tab:edit distance}
    \vspace{-6ex}
\end{table}

\subsection{Quantitative Results}
We compare the performance of the proposed model on two datasets with several baseline methods. Constant prediction model is a simple baseline using the last frame of the past encoding head to perform the prediction; its performance suffers in the transition areas. We also employ Hidden-Markov Model (HMM) as another baseline. Once the past phase encoding likelihoods are obtained, they are used as the observation for HMM. Baum-Welch algorithm \cite{welch2003hidden} is used to estimate the HMM internal parameters. We also compared to a variant of the proposed model, which is the GAN generator trained with only data loss and past encoding loss, which is referred to as Ours w/o Dis. in Table \ref{tab:mgh200}. The experimental results on Cholec80 dataset are shown in Table \ref{tab:Cholec80} and the results on MGH200 are shown in Table \ref{tab:mgh200}. The SUPR-GAN has achieved best overall per-transition accuracy on both datasets\rev{, as measured by the same variety loss used for training. In addition to the prediction module, the proposed SUPR-GAN also contains a past phase recognition module, the past phase recognition accuracy is 82.3\%.} For future predictions, we further show the average Levenshtein distance between the predicted sample and the ground truth, The constant model is having a good performance measured with Levenshtein distance, especially when calculating over all the samples. This is mainly due to the following reasons: (i) when there's no transitions, the constant model makes the reasonable predictions; (ii) For the segments with transition areas, the constant model can still give correct prediction before the actual transition happens. However, after the transition occurs the prediction accuracy can drop significantly, which can be further illustrated by the low per-transition accuracy in Table \ref{tab:mgh200}. Compared to other methods, the proposed model has achieved good performance by having a high per-transition accuracy and maintaining a low Levenshtein distance. 

\rev{Visualizing the predictions, the discriminator encourages a diversity of predictions, while keeping the predictions reasonable. In certain operative phases the surgeon may have freedom of choice on their next phase (e.g. clipping of cystic artery first or duct first or further dissection of the hepatocystic triangle). The discriminator encourages such alternate approaches, resulting in a higher accuracy. However, sequences,  where steps are more deterministic and unique ((e.g. remove GB from liverbed), the discriminator may introduce some noise.} \rev{We note the relative small average performance increase when adding the discriminator compared variety-loss only training. This phenomena has been observed in other prediction tasks \cite{gupta2018social,thiede2019analyzing}. Yet the increase is still significant -- the change in LD gave a p-value of $0.029$ using a pairwise t-test on the per-video scores.}

\subsection{Influence of the Sequence Horizons}
\label{sec:ablation}
The topic of past and future horizon length is inherent to problem formulations in GAN-based temporal analysis \cite{brophy2021generative}
We evaluate the performance of the proposed method with different prediction lengths, as displayed in TABLE \ref{tab:prediction length}. Since LD distance is proportional to the prediction length, we normalize it to the length of 15 seconds. The table shows, that with increased prediction horizon the prediction performance drops, from $2.99$ with prediction horizon of 10 seconds to $7.8$ with prediction horizon of 45 seconds. However, this does not decrease significantly at the clinically chosen prediction horizon of 15 seconds (see Sec I). We also note the degradation of performance with fewer past frames supplied to the model, as the model lacks context information. 
\begin{table}[h!]
\resizebox{.48\textwidth}{!}{
\centering
\begin{tabular}{c|c|c|c|c|c}
\hline
Past steps & Future steps & CM & HMM & Ours w/o Dis. &Ours\\
\hline
  5 & 15 & 11.6 & 14.81&  5.48 &   4.28\\
 10 & 15 & 7.89 & 9.37 & 4.61 & 3.82 \\
 15 & 10 & 7.15 & 13.07&  3.9 & 2.99 \\
 15 & 15 & 3.47 & 13.15 & 4.27 & 3.36 \\
 15 & 30 & 9.24 & 13.31 & 5.31 & 3.95 \\
 15 & 45 & 16.53&  13.9 &5.51 & 7.8 \\
\hline
\end{tabular}
}
\caption{The normalized LD of the different future/past horizons on MGH200 dataset (lower LD is better). As the prediction horizon increases or the past horizon is reduced, the performance of the model drops.} 
\label{tab:prediction length}
   \vspace{-5ex}
\end{table}

\subsection{Surgeon Survey on Phase Identification and Prediction}

To augment the objective measures used and verify the realism of the model's phase prediction, we conducted a survey among a total of 16 surgeons of different educational stages comparing the surgeons' and model prediction. The participants were each presented with 5 out of 20 randomly selected video segments of a LC, 15 seconds each, where the video was stopped at a certain point. The participants were asked to identify the currently performed surgical step. In addition, they were asked to estimate the time remaining for that surgical task until the surgeon proceeds to the next upcoming step. Furthermore participants were asked to evaluate three different possible future trajectories from said point and rank them in terms of likeliness and plausibility to predict the next upcoming surgical step. Two of the presented future trajectories were \rev{chosen uniformly from trajectories} generated by the model whilst one presented the ground truth -- in a broad sense, forming a variant of an \emph{imitation game} \cite{turing2009computing}, as participants distinguish between a real and a computationally-created trajectory. 

\begin{figure}[h!]
    \centering
    \includegraphics[width=.18\textwidth]{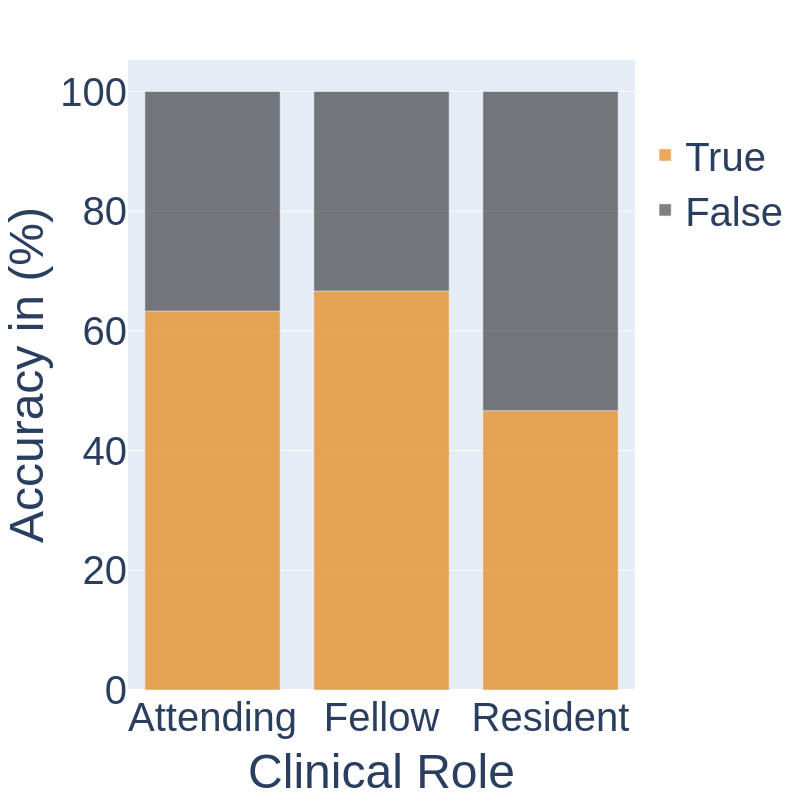}
    \vspace{-0.2cm}
    \caption{Surgeon survey: percentage of correct and incorrect phase prediction according to clinical role and experience.}
    \label{fig:clinical_role}
    \vspace{-0.4cm}
\end{figure}

Firstly to note that the surgeons’ future trajectory classification accuracy varied among participating surgeons and was associated with professional and educational level. Overall, $36,67\%$ of surgical faculty, $33.33\%$ of fellows and $53,33\%$ of residents incorrectly predicted the future surgical step. Overall  only $53.3\%$ of questionnaire answers correctly predicted the future operative step according to the previously determined "ground truth" (see Fig. ~\ref{fig:clinical_role}). Furthermore, the surgeons’ estimation of time until the next operative step displayed a high variance, not linked to experience or type of step, indicating that the estimation of remaining time for future tasks is obviously not a trivial exercise. 

\begin{figure}[h]
\centering
\includegraphics[width=0.45\textwidth]{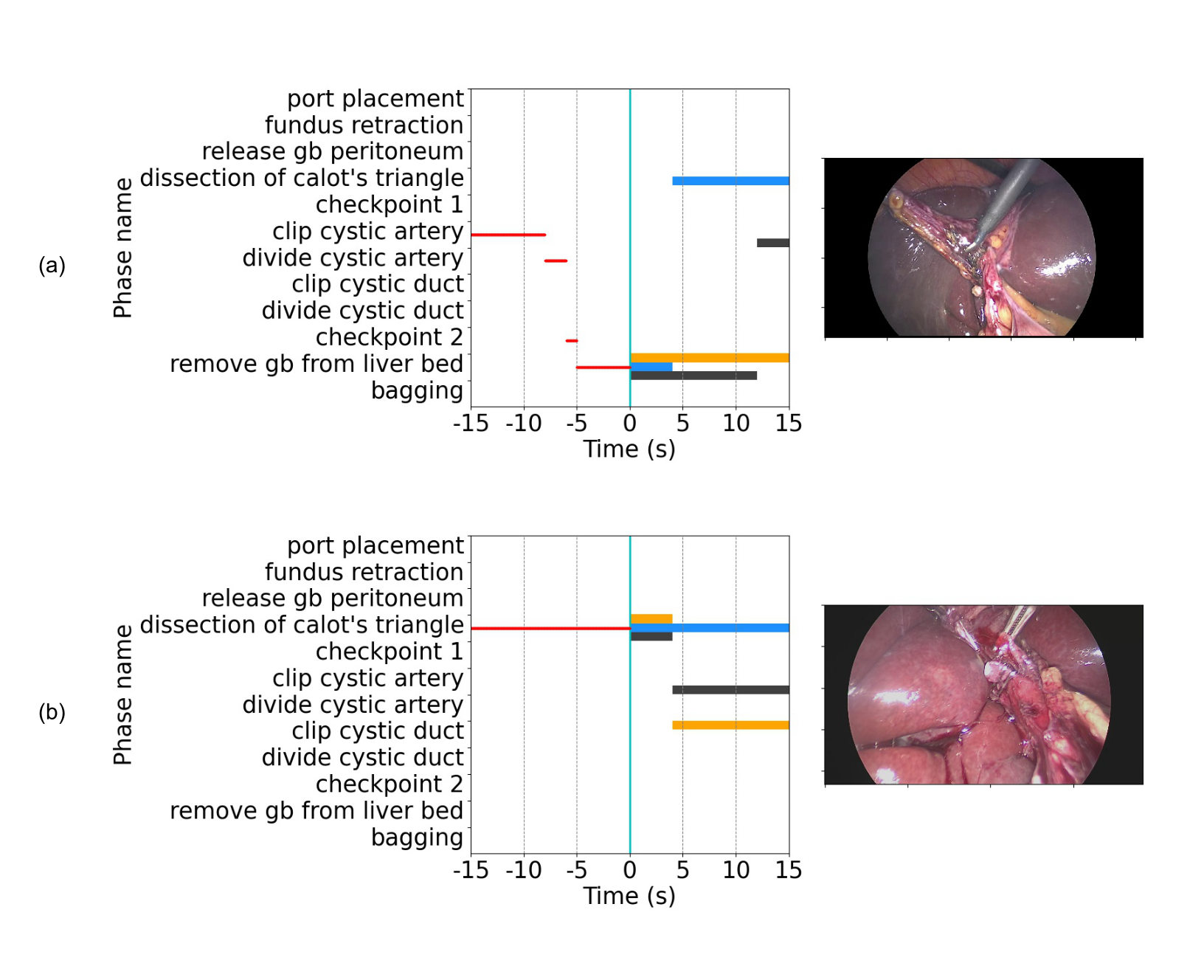}
    \vspace{-0.2cm}
\caption{Examples in the survey. (a) A scenario with unrealistic predicted trajectories -- Yellow represents the ground truth, blue and black show the model predictions. The predictions are unrealistic in the sense that the temporal order of the phases would not occur in an actual surgery. (b) Model proposed trajectories in surgeon survey illustrating alternative future surgical paths. -- Blue represents the ground truth, black and yellow show the model predictions.}
\label{fig:survey_example}
    \vspace{-0.5cm}
\end{figure}

Upon a manual inspection of the 20 presented videos and predicted trajectories by a surgeon, some divergence between the model trajectories and ground truth was observed; but only one model trajectory appeared to be entirely unreasonable, displaying impossible surgical steps. This can be seen in Fig. \ref{fig:survey_example} (a) , where yellow (removal of the gallbladder from the liver bed) represents the ground truth, the model proposed trajectories display surgical steps that chronologically would be performed before this (clipping and dissection). We note that in this scenario our diversity-enhancing loss term resulted in heterogenous samples even though the posterior distribution should not have been diverse, and this can be mitigated by better accuracy-coverage trade-offs. 

In most videos, the model proposed trajectories differing from the ground truth represented feasible alternative surgical approaches. An example can be seen in Fig. \ref{fig:survey_example} (b): the ground truth is displayed by the blue trajectory (dissection of calot’s triangle), although the yellow (clip cystic duct) and black (clip cystic artery) differ entirely from the ground truth, they still represent plausible courses of surgical action. Here only $17\%$ of surgeons selected the correct answer in blue, which is consistent with the heterogenity of surgeons selections in other cases of multiple plausible future steps of the procedure. This shows that the model is not only able to rule out unrealistic future trajectories, but also display multiple alternative, plausible approaches. Overall the model predictions and surgeons' answers aligned well with each other with an accuracy of $53.5\%$ in SUPR-GAN prediction and $53.3\%$ surgeons' choice of suspected ground truth. 

\vspace{-0.1cm}
\section{Conclusions}
\vspace{-0.1cm}
In this paper, we proposed a prediction framework for surgical workflow based on a discrete encoder-decoder GAN to foresee surgical phases in laparoscopic cholecystectomy. Our evaluation on objective metrics, alongside the results of the performed survey of perceived plausibility demonstrate the effectiveness and reliability of the approach. Furthermore, our research suggests several possible opportunities to extend this approach to additional predictive tasks and more exhaustive exploration of video-based surgical process prediction and its relation to the surgical mindset as an statistical inference problem of both analytical and practical value. Due to broad extendability to diverse and more complex surgical procedures, this model holds the potential to significantly enhance surgical decision making and augment risk mitigation for ultimately improved patient outcome. \\
\noindent \textbf{Acknowledgement}
This work uses de-identified surgical videos for model learning. Mass General Brigham IRB (protocol number 2018P002340) approved secondary use of surgical data for this project.
\vspace{-0.3cm}
\bibliographystyle{abbrv}
\bibliography{ref}
\end{document}